\documentclass[pdflatex,sn-mathphys-num]{sn-jnl}% Math and Physical Sciences Numbered Reference Style 
\usepackage{graphicx}%
\usepackage{multirow}%
\usepackage{amsmath,amssymb,amsfonts}%
\usepackage{amsthm}%
\usepackage{mathrsfs}%
\usepackage[title]{appendix}%
\usepackage{xcolor}%
\usepackage{textcomp}%
\usepackage{manyfoot}%
\usepackage{booktabs}%
\usepackage{algorithm}%
\usepackage{algorithmicx}%
\usepackage{algpseudocode}%
\usepackage{listings}%
\usepackage{array}
\usepackage{mathtools}
\usepackage{nth}
\usepackage{natbib}
\newcolumntype{C}[1]{>{\centering\arraybackslash}m{#1}} 
\newcolumntype{R}[1]{>{\raggedright\arraybackslash}m{#1}}

\theoremstyle{thmstyleone}%
\theoremstyle{thmstyletwo}%

\theoremstyle{thmstylethree}%

\begin{document}

\title[Article Title]{RSDiff: Remote Sensing Image Generation from Text using Diffusion Model}

%%=============================================================%%
%% GivenName	-> \fnm{Joergen W.}
%% Particle	-> \spfx{van der} -> surname prefix
%% FamilyName	-> \sur{Ploeg}
%% Suffix	-> \sfx{IV}
%% \author*[1,2]{\fnm{Joergen W.} \spfx{van der} \sur{Ploeg} 
%%  \sfx{IV}}\email{iauthor@gmail.com}
%%=============================================================%%

\author*[1]{\fnm{Ahmad} \sur{Sebaq}}\email{a.sebaq@nu.edu.eg}
\author[1]{\fnm{Mohamed} \sur{ElHelw}}\email{melhelw@nu.edu.eg}
\affil[1]{\orgdiv{Center for Informatics Science (CIS), School of Information Technology and Computer Science (ITCS)}, \orgname{Nile University}, \orgaddress{\street{26th of July Corridor}, \city{First Al Sheikh Zayed}, \postcode{12677}, \state{Giza}, \country{Egypt}}}

%%==================================%%
%% Sample for unstructured abstract %%
%%==================================%%

\abstract{The generation and enhancement of satellite imagery are critical in remote sensing, requiring high-quality, detailed images for accurate analysis. This research introduces a two-stage diffusion model methodology for synthesizing high-resolution satellite images from textual prompts. The pipeline comprises a Low-Resolution Diffusion Model (LRDM) that generates initial images based on text inputs and a Super-Resolution Diffusion Model (SRDM) that refines these images into high-resolution outputs. The LRDM merges text and image embeddings within a shared latent space, capturing essential scene content and structure. The SRDM then enhances these images, focusing on spatial features and visual clarity. Experiments conducted using the Remote Sensing Image Captioning Dataset (RSICD) demonstrate that our method outperforms existing models, producing satellite images with accurate geographical details and improved spatial resolution.}

\keywords{diffusion models; text-to-image generation, remote sensing, generative models, super-resolution}

%%\pacs[JEL Classification]{D8, H51}

%%\pacs[MSC Classification]{35A01, 65L10, 65L12, 65L20, 65L70}

\maketitle

\section{Introduction}\label{sec1}
\label{sec:intro}
Satellite imagery is crucial in various domains, including remote sensing, climate monitoring, and urban planning \cite{ghamisi2017advanced}. The ability to generate high-quality satellite images from text prompts has significant implications for data augmentation, simulation, and enhancing the accessibility of satellite data in resource-constrained environments \cite{xu2022universal, zhang2022artificial}. Traditional methods for generating satellite images often rely on convolutional neural networks (CNNs) \cite{sermanet2012convolutional} or generative adversarial networks (GANs) \cite{goodfellow2014generative}, but they demand large datasets and considerable computational resources \cite{chen2021remote, bejiga2019retro, zhao2021text}.

Diffusion models \cite{ho2020denoising} offer several benefits in the realm of data generation. Firstly, they provide a powerful framework for modeling complex distributions in high-dimensional spaces. By iteratively applying a diffusion process, these models can capture intricate dependencies and generate realistic samples that align with the true underlying data distribution. This capability makes diffusion models well-suited for tasks such as data augmentation, where diverse and plausible samples are required to enhance the training process of deep learning models.

In the domain of image super-resolution, diffusion models also bring noteworthy advantages \cite{ho2022cascaded}. One key strength lies in their ability to generate high-quality images at different resolutions through a sequential process. By starting with a lower-resolution image and gradually refining it, diffusion models excel at capturing global dependencies and preserving fine details. This step-wise approach allows for efficient processing and facilitates the modeling of realistic textures and structures in real-world images. Moreover, using large-scale datasets in the training process of diffusion models enables the acquisition of prior knowledge that aids in producing visually appealing and faithful super-resolved images.

In this paper, a pioneering approach is presented that leverages the power of diffusion models to generate satellite images from text prompts. A novel pipeline is proposed, composed of two diffusion models: an LRDM and an SRDM. The LRDM creates low-resolution satellite images based on textual prompts, capturing the fundamental content and layout of the targeted scenes. The SRDM then takes these low-resolution images as input and super-resolves them, infusing fine-grained spatial details and enhancing visual fidelity to produce high-resolution satellite images. Notably, this dual diffusion model approach operates with only 0.75 billion parameters, which is relatively small compared to other diffusion-based models and generative models in general. This enables us to generate realistic and visually compelling satellite imagery aligned with the specified textual descriptions.

The main contributions of this work are as follows:
\begin{itemize}
    \item A novel pipeline is proposed that combines two diffusion models, allowing us to generate high-resolution satellite images from text prompts efficiently. The stepwise generation process ensures better control over image synthesis and enhanced spatial precision. 
    \item Experiments show that our diffusion pipeline performs better on satellite image synthesis. It achieves the new Fréchet Inception Distance (FID) SoTA for image synthesis with only $\sim0.75$ billion parameters in total.
\end{itemize}

\section{Related Work}
\label{sec:related_work}
\subsection{Generative Adversarial Networks}
The challenge of generating images from textual descriptions is a significant and complex undertaking, with the objective of producing realistic visuals based on natural language prompts. The initial investigations into text-to-image generation predominantly center around algorithms based on GANs. The initial groundbreaking research involves the utilization of a text-conditional GAN \cite{chen2021remote}. Within this framework, the generator is meticulously designed to generate realistic images by harnessing extracted text features and outsmarting the discriminator. Conversely, the discriminator's objective is to accurately distinguish whether the input image is authentic or fabricated. Nevertheless, this particular approach is limited to producing images that possess a spatial dimension of 128x128. 

In order to enhance the spatial resolution of generated images, important point locations were one of the supplementary inputs that scholars have explored for inclusion in the generator. Reed et al \cite{reed2016learning}. proposed the utilization of the generative adversarial what-where network as a means to synthesize images with a spatial scale of 128x128. Another notable contribution is the StackGAN framework \cite{zhang2017stackgan}, which introduces the concept of synthesizing images using a stacked generator, as proposed by the authors. Although the initial generator produces images of limited quality, measuring just 64x64 in spatial scale, the subsequent generator will utilize these images as input and enhance their synthesis to a larger dimension of 256x256.

\subsection{Diffusion Probabilistic Models}
In addition to the utilization of GAN-based techniques, current studies have placed emphasis on the exploration of transformer-based models. DALL-E stands out as a significant work in the field, building upon the foundations laid by GPT-3 (Generative Pre-trained Transformer-3) \cite{brown2020language}, with an impressive parameter count of 12 billion \cite{ramesh2021zero}. Due to the extensive training data consisting of 250 million text-image pairs and the remarkable capacity for learning exhibited by the huge transformer model, DALL-E demonstrates the capability to effectively merge disparate linguistic concepts and generate visually impressive images with a spatial resolution of 256 x 256. The publication of the second iteration of DALL-E has occurred in the recent past\cite{ramesh2022hierarchical}. DALL-E 2 exhibits enhanced capabilities in comparison to its predecessor, DALL-E 1, since it employs a more sophisticated artificial intelligence framework. Similar to the CLIP (contrastive language–image pre-training) model\cite{radford2021learning}, DALL-E 2 acquires knowledge by directly assimilating the association between images and textual descriptions. The diffusion model is capable of producing high-quality images, both realistic and artistic in nature, with a spatial resolution of 1024x1024. These images are generated by leveraging the provided text descriptions. Following DALL-E 2 work recent method \cite{chen2024railfod23} takes a unique approach and leverages large-scale models such as ChatGPT (Chat Generative Pre-trained Transformer) and text-to-image generation models, to synthesize a series of foreign object data.

Flow and energy-based models represent another category of techniques utilized in image generation and optimization \cite{chen2024railfod23, 10035427, 8962207}. These models are particularly effective for capturing complex probability distributions and generating high-quality images by learning the underlying data distribution without relying on Markov chains or variational bounds \cite{10538013, 8424453}. In remote sensing, their application could significantly enhance the authenticity and plausibility of generated images, addressing the stringent requirements of remote sensing scenarios. By leveraging the capabilities of flow and energy-based models, researchers can further advance the field, pushing the boundaries of what is possible in generating realistic remote sensing imagery from textual inputs.

Currently, the majority of research efforts have been directed toward the development of natural images. However, the corresponding studies within the remote sensing community remain underdeveloped. The generation of realistic remote sensing images from text descriptions continues to be a problem due to the increased demands for authenticity and plausibility in application scenarios of remote sensing jobs.

\begin{figure*}[h!]
\begin{center}
 \includegraphics[width=0.9\textwidth]{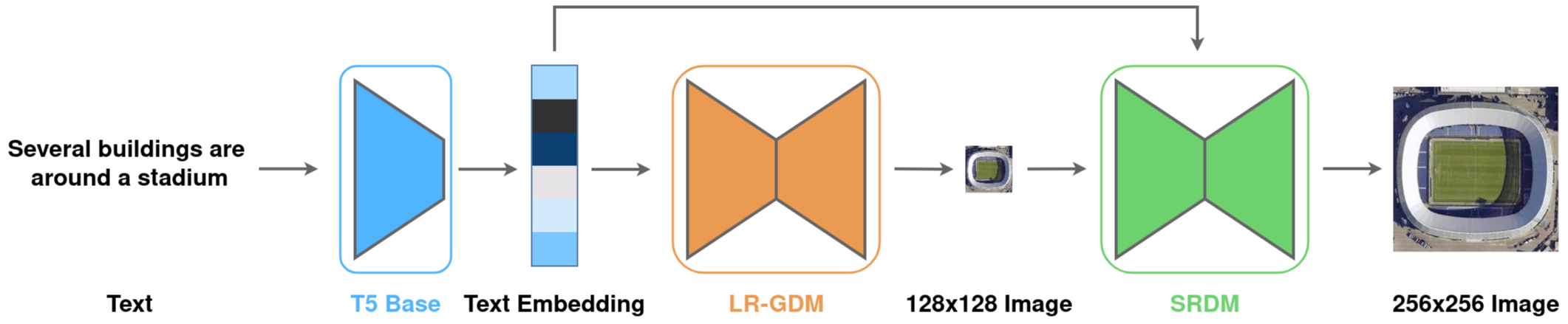}
\end{center}
\caption{The RSDiff framework employs the T5 text encoder to produce text embeddings from the input text. The conditional diffusion model is employed to convert the textual embedding into a 128x128 image. For image upsample, the RSDiff uses a text-conditional super-resolution diffusion model, thereby enhancing the image resolution to a dimension of 256x256 pixels.}
\label{figure:model_arch}
\end{figure*}

\section{Methodology}
\label{sec:method}
The predominant focus of current research in the remote sensing domain pertaining to text-to-image generation lies in the realm of GAN-based methodologies \cite{chen2021remote, bejiga2019retro, zhao2021text}. In contrast, the proposed method centers around a diffusion-based technique. The primary component of our approach is the sequential progression of numerous diffusion processes. The pipeline consists of two diffusion models. Given the considerable success and exceptional output quality of Imagen \cite{saharia2022photorealistic}, A similar technique was adopted in our pipeline. The architecture of our model comprises a text encoder responsible for transforming text into a series of embeddings. Additionally, it incorporates a series of conditional diffusion models that progressively convert these embeddings into images with higher resolutions. 

\subsection{Pretrained text encoders}
Language models \cite{devlin2018bert, raffel2020exploring} are trained on a corpus consisting solely of text, which is notably bigger in size compared to paired image-text data. As a result, these models are exposed to a vast and diverse distribution of textual information. The size of these models is typically larger than the text encoders seen in existing image-text models. Therefore, it is logical to investigate both categories of text encoders for the task of text-to-image conversion. Pretrained text encoder T5 \cite{raffel2020exploring, raffel2017online} is utilized in our approach as it uses a unified framework where all tasks, both generation and classification, are treated as text-to-text problems. Furthermore, T5 exhibits a hybrid nature, being trained to produce either single words or multiple words for a single mask. This flexibility enables the model to effectively learn the intricate structure of language. To ensure simplicity, the weights of the text encoder are held constant. The process of freezing offers numerous benefits, including the ability to do offline computation of embeddings. This feature leads to minimal computational requirements and memory use for training the text-to-image model. It is evident that increasing the size of the text encoder enhances the quality of text-to-image generation \cite{saharia2022photorealistic}. 

\subsection{Diffusion models and classifier-free guidance}
Diffusion models \cite{ho2020denoising, sohl2015deep, song2019generative} are a subset of generative models that employ an iterative denoising technique to convert Gaussian noise into samples originating from a learned data distribution. The models possess the capacity to exhibit conditionality, wherein they can rely on many factors such as class labels, textual inputs, or low-resolution images \cite{dhariwal2021diffusion, ho2022cascaded, nichol2021glide, ramesh2022hierarchical, saharia2022palette, saharia2022image, whang2022deblurring}.

The diffusion models consist of two main processes, a forward process (or diffusion process), in which an image is progressively noised. The variance schedule needs to be defined, with noise levels typically increasing during the forward process; these levels are set as time-dependent constants, although they can be learned. A linear schedule from $\beta_1 = 10^{-4}$ to $\beta_T = 0.02$ is chosen, which gradually adds Gaussian noise to the image. The reverse process (or reverse diffusion process), in which noise is transformed back into a sample from the target distribution. The reverse process is actually a reverse Markov transitions \cite{ho2020denoising}

The training of a diffusion model $\hat{\mathbf{x}}_\theta$ involves the utilization of a denoising objective, which has the following form:
\begin{equation}
    \mathbb{E}_{\mathbf{x},\mathbf{c},\mathbf{\epsilon},t}[w_t||\hat{\mathbf{x}}_\theta(\alpha_t\mathbf{x}+\sigma_t\mathbf{\epsilon},\mathbf{c})-\mathbf{x}||^2_2]
    \label{eq:diffusion}
\end{equation}
Where $(\mathbf{x}, \mathbf{c})$ represent pairs of data and their corresponding conditioning factors, $t\sim\mathcal{U}([0,1]), \mathbf{\epsilon\sim\mathcal{N}(\mathbf{0},\mathbf{I})}$, and $\alpha_t, \sigma_t, w_t$ are functions that impact the quality of the sample. $\hat{\mathbf{x}}_\theta$ is trained in a manner that aims to remove noise from $\mathbf{z}_t \coloneqq \alpha_t\mathbf{x}+\sigma_t\epsilon$ and produce $\mathbf{x}$, utilizing a squared error loss function that is weighted to prioritize specific values of $t$. Ho et al. \cite{ho2020denoising} observed that the simple mean-sqaured error objective works better in practice.

The utilization of classifier guidance \cite{dhariwal2021diffusion} is a method employed to enhance the quality of samples and decrease variability in conditional diffusion models. This is achieved by including gradients derived from a pre-trained model, namely $p(c|z_t)$, during the sampling process. The utilization of classifier-free guidance \cite{ho2022classifier} approach presents an alternate methodology that circumvents the reliance on a pre-trained model. Instead, it involves the simultaneous training of a solitary diffusion model on both conditional and unconditional objectives. This is achieved by randomly deleting a specific parameter, denoted as $c$ during the training process, typically with a chance of 10\%. The process of sampling is carried out with the adjusted x-prediction technique:
\begin{equation}
    \hat{\epsilon}_\theta(\mathbf{z}_t,\mathbf{c})=w\epsilon_\theta(\mathbf{z}_t,\mathbf{c})+(1-w)\epsilon_\theta(\mathbf{z}_t)
    \label{eq:cfg}
\end{equation}
Here, $\hat{\epsilon}_\theta(\mathbf{z}_t,\mathbf{c})$ is the final prediction for the noise model used to reverse the diffusion process at time step $t$. It's computed as a weighted combination of conditional and unconditional predictions of the noise. $\epsilon_\theta(\mathbf{z}_t,\mathbf{c})$ represents the conditional prediction of the noise vector, where the condition is a text description that guides the generation. It's based on the current state of the latent variable $\mathbf{z}_t$ and the condition $\mathbf{c}$. $\epsilon_\theta(\mathbf{z}_t)$  is the unconditional prediction of the noise. It only depends on the latent variable $\mathbf{z}_t$ and is used when no specific guidance (text descriptions) is provided. $w$ is the guidance weight, it controls the contribution of the conditional versus unconditional noise predictions in the final noise model. When $w$ is set to 1, the model effectively ignores the conditional inputs (classifier-free guidance is disabled). When increasing the value of $w$ above 1 strengthens the influence of the conditional component $\epsilon_\theta(\mathbf{z}_t,\mathbf{c})$ enhancing the model's responsiveness to the guiding condition $\mathbf{c}$. The effectiveness of text conditioning in RSDiff relies heavily on the utilization of classifier-free advice.

\subsection{Cascaded diffusion models}
The proposed methodology involves employing a sequential process consisting of a foundational 128×128 model, alongside text-conditional super-resolution diffusion models, to enhance the resolution of a generated image from 128 x 128 to 256 x 256 (see Figure \ref{figure:model_arch}). The utilization of cascaded diffusion models, together with noise conditioning augmentation, has proven to be highly successful in the gradual generation of images with a high level of fidelity \cite{ho2022cascaded}. 
The authors of \cite{ho2022cascaded} utilize three models: one for image generation and two for super-resolution, while our approach employs a two-stage process—one for generation and another for super-resolution—achieving the same final resolution of 255x255. The T5 model was employed to facilitate text-based conditional generation. Additionally, whereas \cite{ho2022cascaded} requires 4000 iterations of a diffusion process to generate an image, our method needs fewer than 1000 iterations. Our experiments have demonstrated that using a generation model capable of directly producing a 128x128 image, which is then super-resolved to 256x256, eliminates the need for an intermediate super-resolution model. Consequently, comparable was achieved results with fewer models, fewer steps, and reduced computational requirements. Furthermore, the integration of noise level conditioning into the super-resolution models not only improves the quality of the generated samples but also enhances the robustness of these models in dealing with artifacts generated by lower-resolution models \cite{ho2022cascaded}. The utilization of noise conditioning augmentation is implemented in both of the super-resolution models. The integration of noise level conditioning into the super-resolution models is crucial for creating images with high precision and complexity, as it effectively manages corruption introduced by augmentation based on specific conditioning levels. Subsequently, the diffusion model is conditioned on the augmented image. During the training phase, the amount of augmentation is selected in a random manner. However, during the inference phase, various values of augmentation were systematically explored in order to identify the optimal sample quality. Gaussian noise is used as an augmentation method that utilizes variance-preserving Gaussian noise augmentation that replicates the forward process employed in diffusion models.

\subsection{Neural network architecture}
The U-Net architecture \cite{nichol2021improved} has been adapted for the underlying 128x128 text-to-image diffusion model. Text embeddings are integrated using a pooled embedding vector, which is merged with the diffusion timestep embedding in a manner similar to the class embedding conditioning technique \cite{dhariwal2021diffusion, ho2022cascaded}. To enhance the model while maintaining its lightness, only 4 upsampling and downsampling stages were used. Cross-attention across the text embeddings was incorporated at the last 3 stages by conditioning on the complete sequence of text embeddings. Additionally, Layer Normalisation was employed in the attention and pooling layers of text embeddings, which significantly enhanced performance. For upsampling and downsampling activations, only 3 residual blocks \cite{song2020score} were used.

The Efficient U-Net model \cite{saharia2022photorealistic} is employed for achieving super-resolution from 128x128 to 256x256. Its improvements were leveraged to enhance memory efficiency, reduce inference time, and improve convergence speed, with the self-attention layer used only in the last stage of upsampling and downsampling. The last text cross-attention layers were retained, as they were found to be critical.

A decision was made to generate a 128x128 image and then super-resolve it to 256x256 instead of generating a 256x256 image in one step, influenced by several factors:
\begin{itemize}
    \item \textbf{Computational efficiency:} Diffusion models can be computationally intensive, especially when applied to high-resolution images. By initially generating a lower-resolution image, the computational burden was alleviated, resulting in faster processing and training times. This approach enables more practical implementation and facilitates experimentation with larger datasets.
    \item \textbf{Enhanced information flow:} Diffusion models often operate in an autoregressive manner, processing a sequence of lower-resolution images or noise samples. Starting with a lower resolution allows for better information flow and conditioning at each step of the diffusion process. This sequential generation aids in accurately modeling the data distribution and capturing complex dependencies within the image.
    \item \textbf{Capturing global dependencies:} Generating a lower-resolution image initially enables diffusion models to capture global dependencies and structures more effectively. This step allows the model to learn the overarching patterns and relationships present in the image. By utilizing this understanding, the subsequent super-resolution process can ensure that the resulting high-resolution image aligns with the overall structure of the target image.
    \item \textbf{Prioritization of details:} The initial lower-resolution generation in diffusion models allows for the prioritization of important image details and textures during the subsequent super-resolution step. By focusing on recovering these critical aspects, the model can produce enhanced high-resolution images that preserve realistic and fine-grained details. This is particularly advantageous in real-world image super-resolution tasks, where capturing and preserving such details are essential for generating visually pleasing and faithful results.
\end{itemize}

\section{Experiments}
\subsection{Dataset}
The study employed the Remote Sensing Image Captioning Dataset (RSICD) \cite{lu2017exploring}. The RSICD dataset was originally collected with the explicit intention of supporting the photo captioning task in the field of remote sensing. The RSICD was used to evaluate the proposed methodology in this research endeavor. The dataset consists of a thorough compilation of 10,921 remote-sensing photographs characterized by exceptional resolution. Each image in the dataset is characterized by dimensions of 224x224 pixels and is accompanied by five textual descriptions serving as annotations. The collection has 30 unique scenario categories. The experiment employs a training set of 8734 pairs of text and images from the training split. The test set, on the other hand, is comprised of the remaining 2187 combinations of text and images\cite{xu2022txt2img}. In the experiment that was done,all images underwent a resizing process using bilinear interpolation, which yielded a consistent resolution of 256x256 pixels.

\subsection{Evaluation metrics}
The evaluation metrics employed in this study are the Inception Score (IS) \cite{salimans2016improved} and the FID \cite{heusel2017gans}. The IS evaluates a model's ability to represent the entire ImageNet \cite{deng2009imagenet} class distribution while producing samples that convincingly belong to specific classes. However, IS has limitations, such as not rewarding the coverage of the entire distribution or diversity within a class. Models that simply memorize a limited subset of the dataset can still achieve high IS \cite{barratt2018note}. To address these issues and better assess diversity, the FID is used. FID is considered more aligned with human judgment than IS and quantifies the similarity between two image distributions by measuring the distance in the latent space of Inception-V3 \cite{szegedy2016rethinking}.
These metrics are utilized to assess the degree of stylistic resemblance between the original samples and the generated ones \cite{radford2021learning, zhou2111lafite} in the RSICD. 
In this context, the function $g(z)$ represents the image that is formed and needs to be assessed. Additionally, for a specific label $l$ the posterior probability $q(l|x)$ is computed by the Inception-V3 model for a given image $x$. Then, IS can be calculated as:
\begin{equation}
    IS=exp[\mathbb{E}_{z\sim q(z)}[D_{kl}(q(l|g(z))||q(l)]]
    \label{eq:is}
\end{equation}
The marginal class distribution is $q(l)$ and $D_{kl}(.)$ refers to the KL-divergence, which is a measure of the difference between two probability distributions.

The FID score is calculated by utilizing the features of the last average pooling layer within the Inception-V3 model. Then, IS can be calculated as:
\begin{equation}
    FID=||\mu_r - \mu_g||^2 + Tr((\Sigma_r+\Sigma_g-2(\Sigma_r\Sigma_g)^{\frac{1}{2}})
    \label{eq:fid}
\end{equation}
Where $(\mu_g, \Sigma_g)$ and $(\mu_r, \Sigma_r)$ represent the mean and covariance of the generated and real features, respectively. The function $Tr(.)$ denotes the trace operation in linear algebra.

\subsection{Training}
A significantly lighter diffusion model is employed for image synthesis consisting of 260 million parameters compared to the one utilized in Imagen \cite{saharia2022photorealistic} which consists of 2 billion parameters. Additionally, a 260 million parameter model is employed for super-resolution tasks which is also lighter than the one used in Imagen with 600 million parameters. Both models were trained using a batch size of 64 and 1000 training epochs. For the diffusion process, hyperparameter selection was guided by the work of \cite{saharia2022photorealistic, ho2020denoising} as their selection showed good performance, the number of iterations $T$ is set to 1000, $\beta_1 = 10^{-4}$ to $\beta_t = 0.02$. The base 128x128 model in our study utilizes the Tesla V100 SXM2 processor, which is equipped with 32GB of memory. Additionally, the super-resolution variant employs another Tesla V100 chip. The Adafactor optimizer \cite{shazeer2018adafactor} is employed for our standard 128x128 model due to its significantly reduced memory requirements compared to Adam while attaining comparable performance. The Adam optimizer \cite{kingma2014adam} is preferred over Adafactor for super-resolution models due to its superior performance. For both optimizers, a learning rate of $1e-4$ was employed, along with 10,000 linear warm up steps. In order to offer classifier-independent guidance, a joint training approach was employed where the text embeddings for all three models are randomly set to zero with a probability of 10\%. This method enhances the models' robustness and flexibility, enabling them to perform effectively without relying on specific classifiers. Without classifier-free guidance, the model might rely heavily on a separate classifier to interpret the text and guide the image generation process. With classifier-free guidance, the model is trained to sometimes ignore the text embeddings, learning to generate relevant images both with and without explicit text guidance. This dual capability enhances the model's versatility and performance \cite{ho2022classifier, dhariwal2021diffusion, saharia2022photorealistic}.

\begin{table*}[h!]
\begin{center}
\resizebox{\textwidth}{!}{
\begin{tabular}{R{3cm}C{2cm}C{2cm}C{2cm}C{2cm}C{2cm}C{2cm}C{2cm}C{2cm}C{2cm}}
Test prompt & DF-GAN & Attn-GAN & DAE-GAN & Lafite & Txt2Img-MHN (VQVAE) & DALL-E & Txt2Img-MHN (VQGAN) & RSDiff (Ours) & Real\\

In front of the sea is a vast beach & \includegraphics[width=2cm]{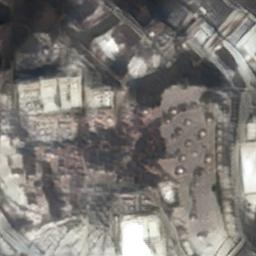} & \includegraphics[width=2cm]{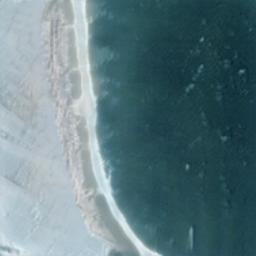} & \includegraphics[width=2cm]{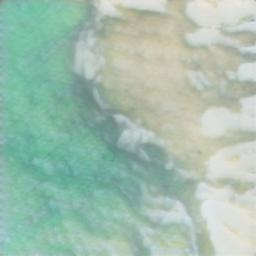} & \includegraphics[width=2cm]{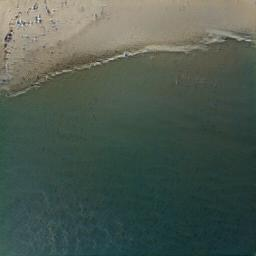} & \includegraphics[width=2cm]{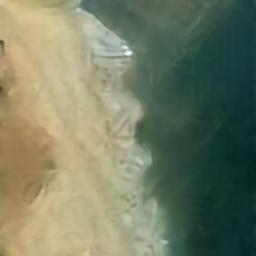} & \includegraphics[width=2cm]{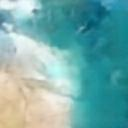} & \includegraphics[width=2cm]{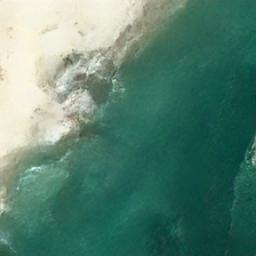} & \includegraphics[width=2cm]{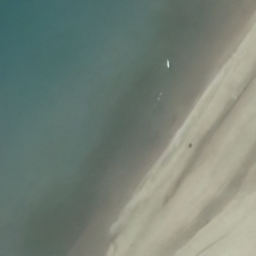} & \includegraphics[width=2cm]{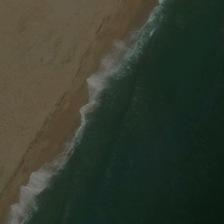} \\

Narrow roads were built around the farm & \includegraphics[width=2cm]{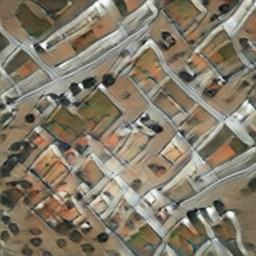} & \includegraphics[width=2cm]{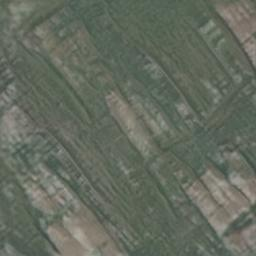} & \includegraphics[width=2cm]{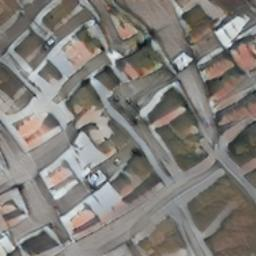} & \includegraphics[width=2cm]{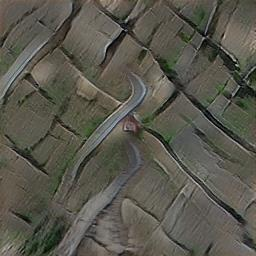} & \includegraphics[width=2cm]{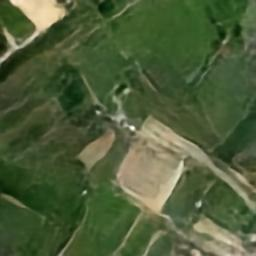} & \includegraphics[width=2cm]{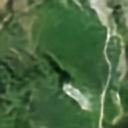} & \includegraphics[width=2cm]{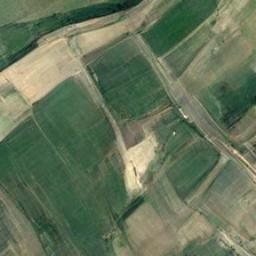} & \includegraphics[width=2cm]{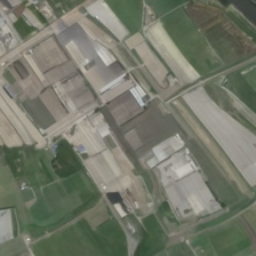} & \includegraphics[width=2cm]{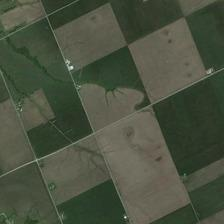} \\

An airport with many buildings beside in it & \includegraphics[width=2cm]{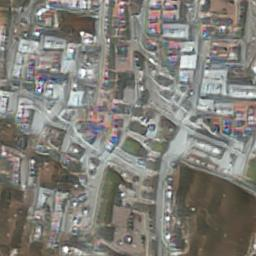} & \includegraphics[width=2cm]{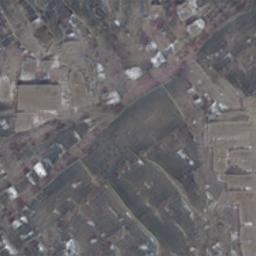} & \includegraphics[width=2cm]{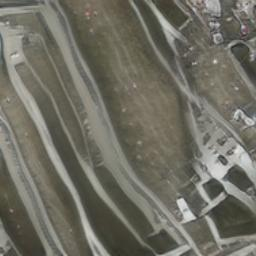} & \includegraphics[width=2cm]{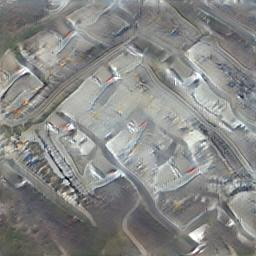} & \includegraphics[width=2cm]{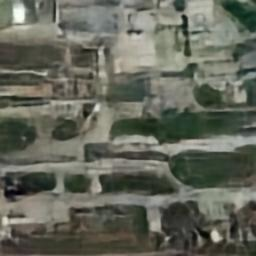} & \includegraphics[width=2cm]{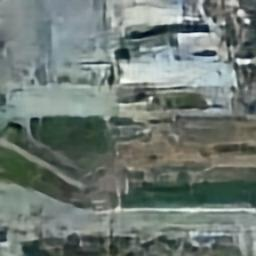} & \includegraphics[width=2cm]{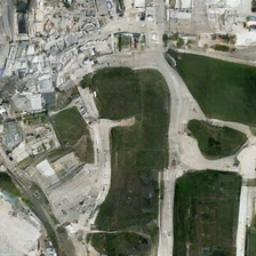} & \includegraphics[width=2cm]{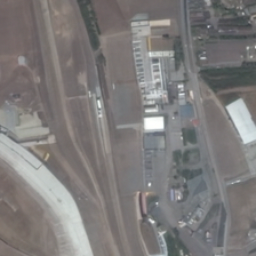} & \includegraphics[width=2cm]{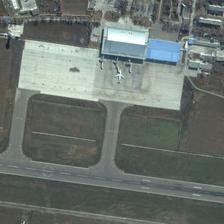} \\

Next to the playground is a house with a grey roof & \includegraphics[width=2cm]{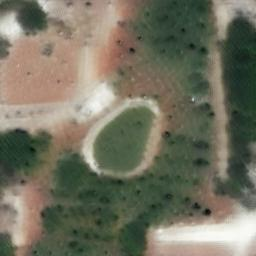} & \includegraphics[width=2cm]{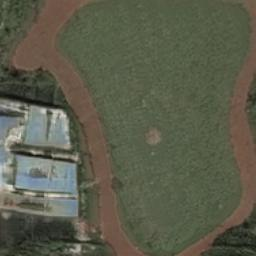} & \includegraphics[width=2cm]{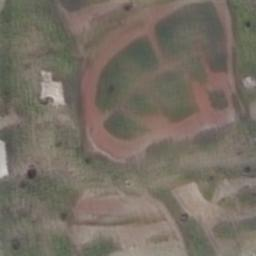} & \includegraphics[width=2cm]{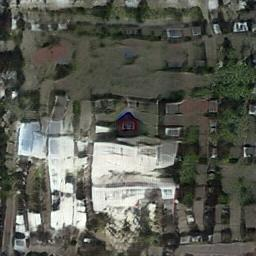} & \includegraphics[width=2cm]{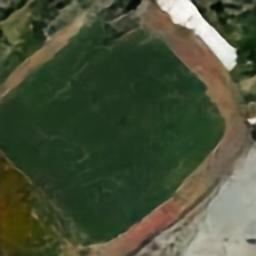} & \includegraphics[width=2cm]{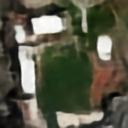} & \includegraphics[width=2cm]{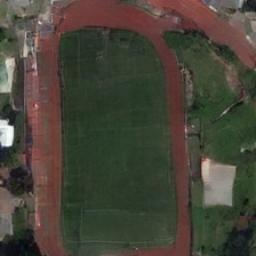} & \includegraphics[width=2cm]{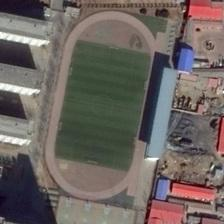} & \includegraphics[width=2cm]{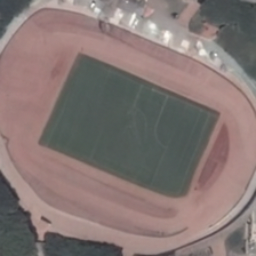} \\

A baseball field is near some green trees & \includegraphics[width=2cm]{1_2.png} & \includegraphics[width=2cm]{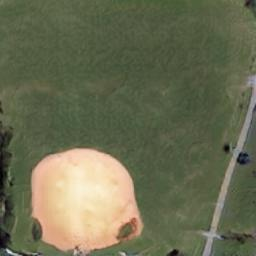} & \includegraphics[width=2cm]{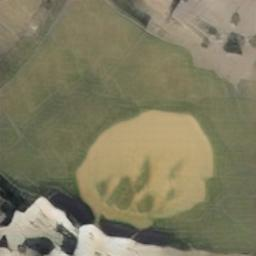} & \includegraphics[width=2cm]{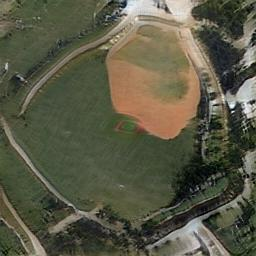} & \includegraphics[width=2cm]{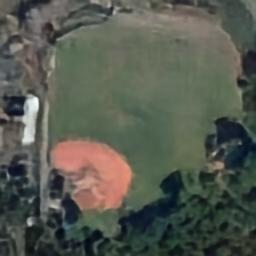} & \includegraphics[width=2cm]{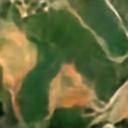} & \includegraphics[width=2cm]{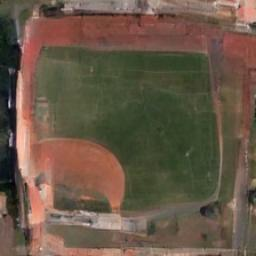} & \includegraphics[width=2cm]{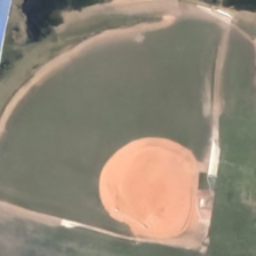} & \includegraphics[width=2cm]{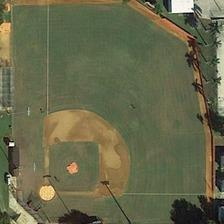} \\

There is no plant on the bare land & \includegraphics[width=2cm]{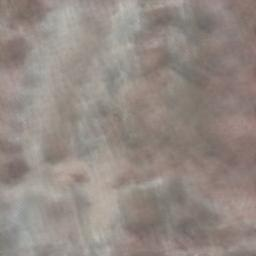} & \includegraphics[width=2cm]{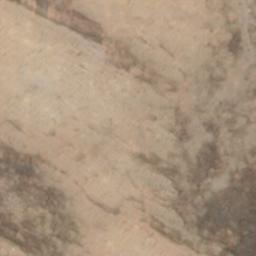} & \includegraphics[width=2cm]{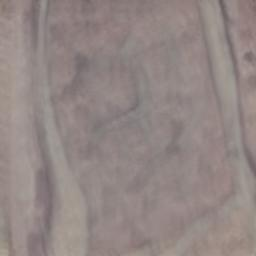} & \includegraphics[width=2cm]{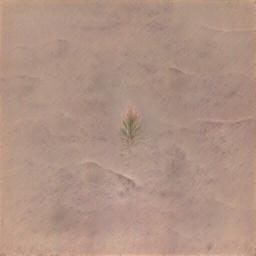} & \includegraphics[width=2cm]{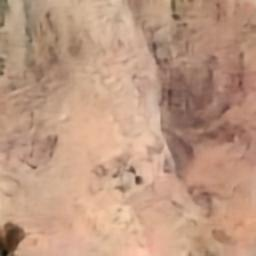} & \includegraphics[width=2cm]{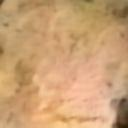} & \includegraphics[width=2cm]{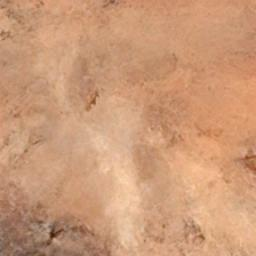} & \includegraphics[width=2cm]{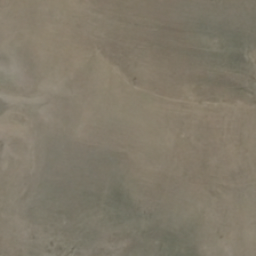} & \includegraphics[width=2cm]{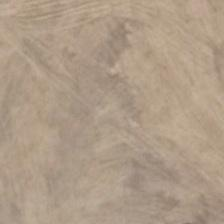} \\

There is a long bridge over the river & \includegraphics[width=2cm]{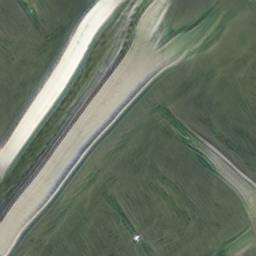} & \includegraphics[width=2cm]{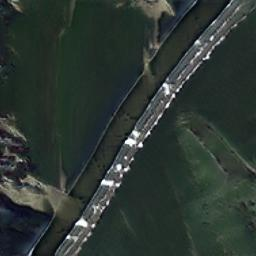} & \includegraphics[width=2cm]{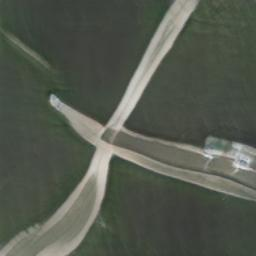} & \includegraphics[width=2cm]{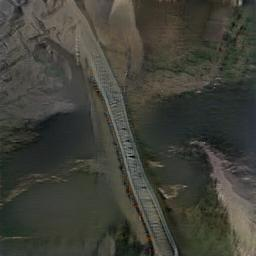} & \includegraphics[width=2cm]{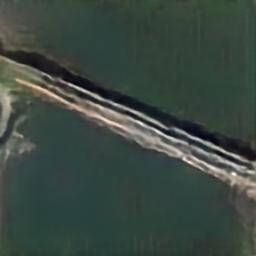} & \includegraphics[width=2cm]{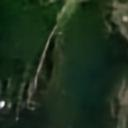} & \includegraphics[width=2cm]{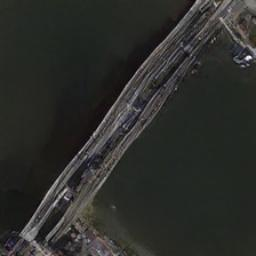} & \includegraphics[width=2cm]{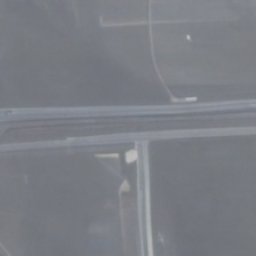} & \includegraphics[width=2cm]{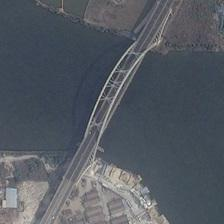} \\

On the ground there are several residential buildings & \includegraphics[width=2cm]{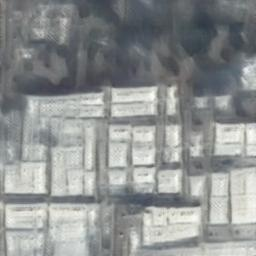} & \includegraphics[width=2cm]{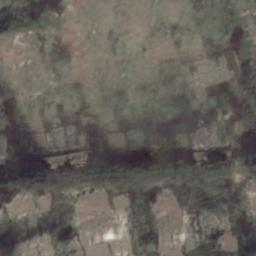} & \includegraphics[width=2cm]{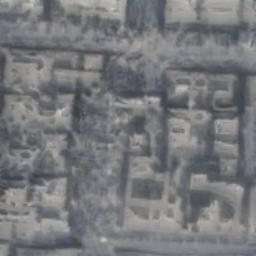} & \includegraphics[width=2cm]{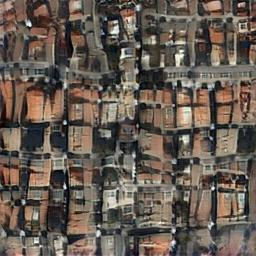} & \includegraphics[width=2cm]{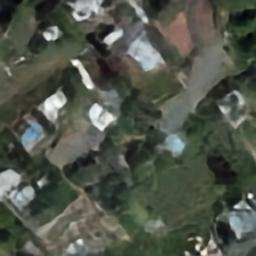} & \includegraphics[width=2cm]{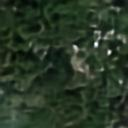} & \includegraphics[width=2cm]{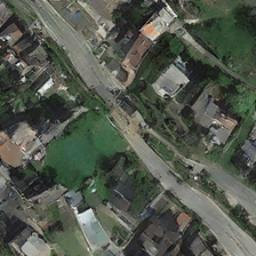} & \includegraphics[width=2cm]{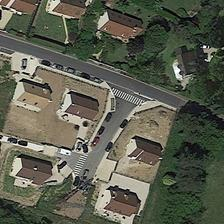} & \includegraphics[width=2cm]{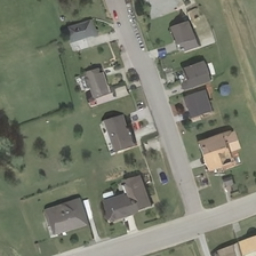} \\

Several buildings are around a stadium & \includegraphics[width=2cm]{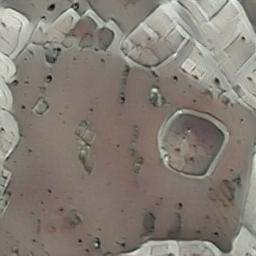} & \includegraphics[width=2cm]{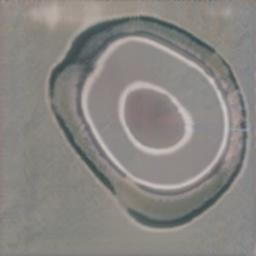} & \includegraphics[width=2cm]{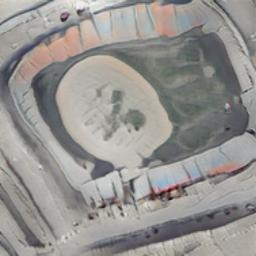} & \includegraphics[width=2cm]{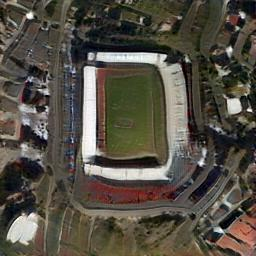} & \includegraphics[width=2cm]{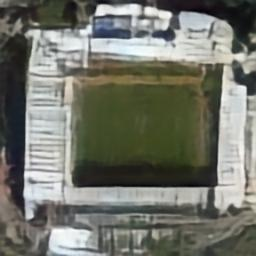} & \includegraphics[width=2cm]{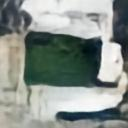} & \includegraphics[width=2cm]{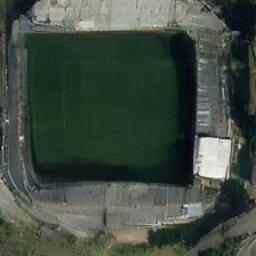} & \includegraphics[width=2cm]{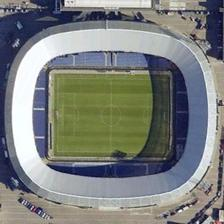} & \includegraphics[width=2cm]{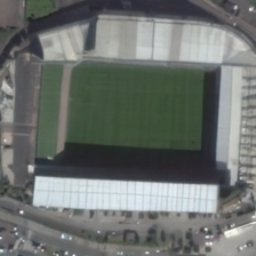} \\

Many buildings are in two sides of a railway station & \includegraphics[width=2cm]{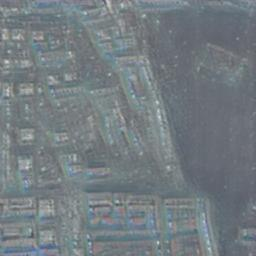} & \includegraphics[width=2cm]{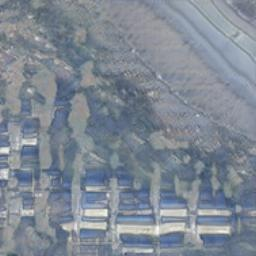} & \includegraphics[width=2cm]{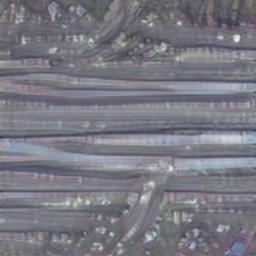} & \includegraphics[width=2cm]{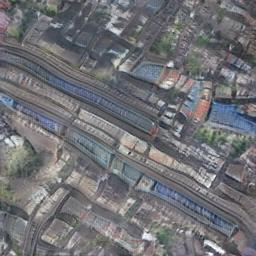} & \includegraphics[width=2cm]{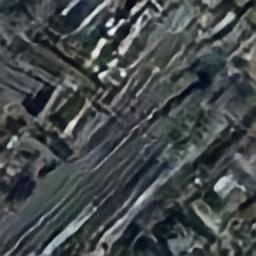} & \includegraphics[width=2cm]{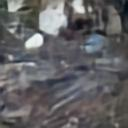} & \includegraphics[width=2cm]{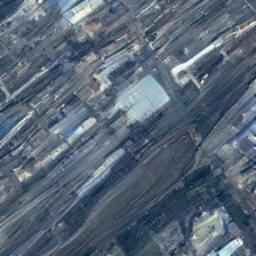} & \includegraphics[width=2cm]{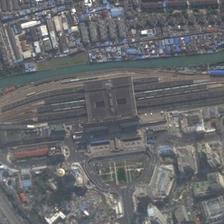} & \includegraphics[width=2cm]{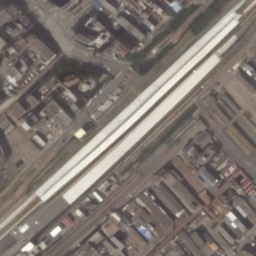} \\

Many buildings and green trees are in a dense residential area & \includegraphics[width=2cm]{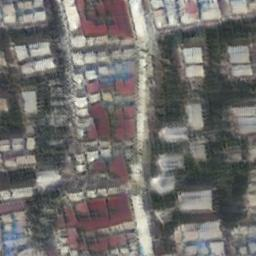} & \includegraphics[width=2cm]{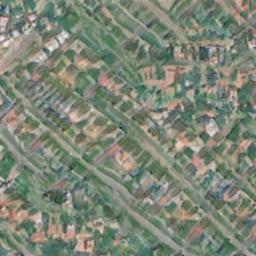} & \includegraphics[width=2cm]{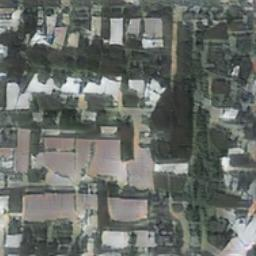} & \includegraphics[width=2cm]{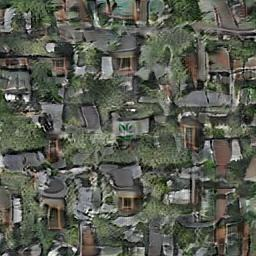} & \includegraphics[width=2cm]{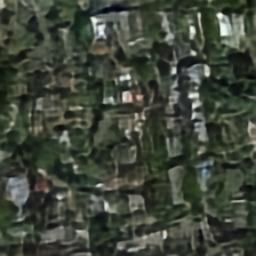} & \includegraphics[width=2cm]{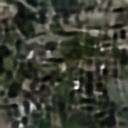} & \includegraphics[width=2cm]{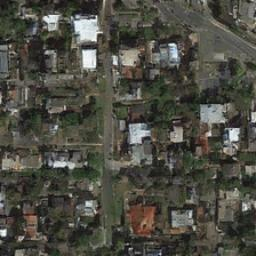} & \includegraphics[width=2cm]{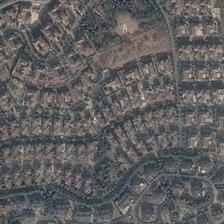} & \includegraphics[width=2cm]{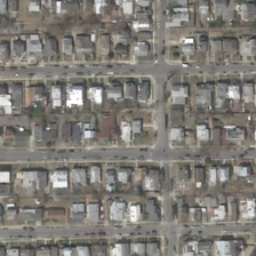} \\

\end{tabular}
}
\end{center}
\caption{Displaying remote sensing images generated through a range of text-to-image generation techniques, all generated from textual descriptions sourced from the test dataset. \cite{xu2022txt2img}}
\label{table:qualitative_results}
\end{table*}

\begin{table*}[ht!]
\begin{center}
\begin{tabular}{cccc}
\toprule
Method & Inception Score $\uparrow$ & FID Score $\downarrow$ \\
\midrule
DF-GAN & 9.51 & 109.41\\
DAE-GAN & 7.71 & 93.15\\
Attn-GAN & \textbf{11.71} & 95.81\\ 
Lafite & 10.70 & 74.11\\
\midrule
DALL-E & 2.59 & 191.93\\
RSDiff (Ours) & 7.22 & \textbf{66.49}\\
Txt2Img-MHN (VQVAE) & 3.51 & 175.36\\
Txt2Img-MHN (VQGAN) & 5.99 & 102.44\\ 
\bottomrule
\end{tabular}
\end{center}
\caption{The quantitative outcomes acquired from the RSICD test dataset. Our study includes the reporting of both the IS and the FID scores.}
\label{table:quantitative_results}
\end{table*}

\section{Results}
Comparing our suggested method to 7 state-of-the-art (SoTA) approaches in text-to-image generation unveils a diverse landscape of innovation in this field. While Attn-GAN integrates attention learning and multi-stage refinement within a GAN framework \cite{xu2018attngan}, DAE-GAN adopts aspect-level synthesis with global and local refinement \cite{ruandae}, and DF-GAN leverages a deep text-image fusion block \cite{tao2022df}. Lafite introduces a language-free paradigm by utilizing the pre-trained CLIP model \cite{zhou2111lafite}, and DALL-E pioneers zero-shot text-to-image generation \cite{ramesh2022hierarchical}. Txt2Img-MHN models (VQVAE and VQGAN) employ a Hopfield network for synthesis \cite{xu2022txt2img}, while our RSDiff model proposes a text-to-image cascaded diffusion models pipeline, offering a novel contribution.

The task of generating remote sensing images that are both photorealistic and semantically consistent is fraught with challenges due to the complexities of spatial distributions. While DF-GAN and other GAN-based techniques excel at replicating visual styles, intricate shapes and boundary details may be challenging. RSDiff demonstrates superior performance in complex scenarios, as evidenced by its ability to capture semantic nuances, such as the varying amounts of described objects, as shown in Table \ref{table:qualitative_results}.

The generation of remote sensing images that are both photo-realistic and semantic-consistent is a highly demanding task, mostly because of the intricate spatial distribution of many ground objects. Although DF-GAN and other advanced GAN-based techniques have demonstrated impressive capabilities in replicating the visual style of remote sensing images, they may encounter challenges when it comes to accurately reproduce intricate shape and boundary details of complex ground targets, such as the playground (located in the \nth{9} row of Table \ref{table:qualitative_results}) and the stadium (located in the \nth{5} row of Table \ref{table:qualitative_results}). In contrast, the suggested RSDiff algorithm has the capability to produce more realistic outcomes in these complex settings.

Furthermore, it has been observed that RSDiff exhibits superior ability in acquiring the notion of amount. Consider the last three rows depicted in Table \ref{table:qualitative_results} as an illustrative example. The three input text descriptions provided by the user pertain to the subject of building objects. However, the number of buildings mentioned in these descriptions varies, ranging from a singular building to many buildings, and finally to a large number of buildings. While the majority of the solutions yield satisfactory outcomes in the initial scenario, comprehending the precise meaning of terms such as "several" and "many" poses a greater challenge. This tendency is most evident in the outcome of Lafite when it erroneously produces a substantial number of densely populated residential structures when presented with textual descriptions of "several buildings." In contrast, the outcomes of RSDiff exhibit a higher degree of semantic consistency with the textual descriptions provided as input.

Assessing the quality of synthesized images typically relies on human observers' visual perception, which is often intuitive. However, accurately quantifying this judgment poses challenges due to the inherent ambiguity and complexity of the semantic information conveyed within the image. Presently, the prevailing metrics employed for image generation encompass the Inception Score and the FID Score \cite{nichol2021glide, ramesh2021zero} Although these two measures are straightforward in nature, they were initially employed to assess the efficacy of the GAN model, with an emphasis on either the resemblance of visual styles or the disparity in features between synthesized and authentic data.

Table \ref{table:quantitative_results} presents a comprehensive overview of the Inception Score and FID Score associated with each approach employed in the present investigation. One intriguing observation is that GAN-based techniques tend to exhibit significantly superior performance on metrics such as the Inception Score and FID Score compared to Transformer-based methods, a finding that aligns with earlier research \cite{ramesh2021zero}. The Attn-GAN, outperforms the proposed RSDiff model, producing an Inception Score of 11.71 compared to RSDiff's 7.22. Simultaneously, it's important to recognize that RSDiff, despite its slightly lower score, holds a notable position as a leading diffusion model. Moreover, pivoting to a different evaluation metric, the suggested methodology shines with remarkable performance in FID score, attaining a state-of-the-art outcome of 66.49.

\section{Conclusion}
Introducing an innovative text-to-image diffusion model, this study significantly advances the generation of remote-sensing images that are not only photorealistic but also semantically aligned with textual descriptions. Our technical approach utilizes a two-stage diffusion process: the first stage generates a basic image outline based on textual inputs, and the second stage enhances these images to high-resolution outputs, ensuring detailed accuracy and visual clarity. A comprehensive evaluation of our model was conducted against current state-of-the-art techniques using the RSICD benchmark dataset. The results from our extensive testing confirm that our model excels in producing remote sensing images with superior authenticity and spatial resolution compared to established methods. This success highlights the efficacy of our dual-stage diffusion process in handling the intricate demands of remote sensing imagery synthesis.

One of the primary challenges that our research addressed is the scarcity of training data specifically paired text-image datasets in remote sensing. While acquiring precise textual descriptions remains a hurdle, leveraging the vast amounts of available unlabeled, high-resolution remote sensing photos presents a promising avenue. Future research will focus on harnessing this semantic data from unlabeled images to further refine and enhance the effectiveness of our text-to-image generation model. Our ongoing and future research endeavors aim to explore these areas, potentially setting new benchmarks in the field.
\section*{Conflict of Interest}
The authors declare that there is no conflict of interest regarding the publication of this paper.

\section*{Data Availability Statement}
The data supporting the findings of this study are openly available in the Remote Sensing Image Captioning Dataset (RSICD), which is hosted by the Remote Sensing Image Captioning Project at their official \href{https://github.com/201528014227051/RSICD_optimal}{repository}. The RSICD is a publicly accessible dataset designed for the development and evaluation of image captioning models in the context of remote sensing imagery. This dataset can be accessed freely by researchers and practitioners in the field, adhering to the terms and conditions specified by the host institution.

%%===========================================================================================%%
%% If you are submitting to one of the Nature Portfolio journals, using the eJP submission   %%
%% system, please include the references within the manuscript file itself. You may do this  %%
%% by copying the reference list from your .bbl file, paste it into the main manuscript .tex %%
%% file, and delete the associated \verb+\bibliography+ commands.                            %%
%%===========================================================================================%%

\bibliography{sn-bibliography}% common bib file

%% BioMed_Central_Bib_Style_v1.01

\begin{thebibliography}{48}
% BibTex style file: bmc-mathphys.bst (version 2.1), 2014-07-24
\ifx \bisbn   \undefined \def \bisbn  #1{ISBN #1}\fi
\ifx \binits  \undefined \def \binits#1{#1}\fi
\ifx \bauthor  \undefined \def \bauthor#1{#1}\fi
\ifx \batitle  \undefined \def \batitle#1{#1}\fi
\ifx \bjtitle  \undefined \def \bjtitle#1{#1}\fi
\ifx \bvolume  \undefined \def \bvolume#1{\textbf{#1}}\fi
\ifx \byear  \undefined \def \byear#1{#1}\fi
\ifx \bissue  \undefined \def \bissue#1{#1}\fi
\ifx \bfpage  \undefined \def \bfpage#1{#1}\fi
\ifx \blpage  \undefined \def \blpage #1{#1}\fi
\ifx \burl  \undefined \def \burl#1{\textsf{#1}}\fi
\ifx \doiurl  \undefined \def \doiurl#1{\url{https://doi.org/#1}}\fi
\ifx \betal  \undefined \def \betal{\textit{et al.}}\fi
\ifx \binstitute  \undefined \def \binstitute#1{#1}\fi
\ifx \binstitutionaled  \undefined \def \binstitutionaled#1{#1}\fi
\ifx \bctitle  \undefined \def \bctitle#1{#1}\fi
\ifx \beditor  \undefined \def \beditor#1{#1}\fi
\ifx \bpublisher  \undefined \def \bpublisher#1{#1}\fi
\ifx \bbtitle  \undefined \def \bbtitle#1{#1}\fi
\ifx \bedition  \undefined \def \bedition#1{#1}\fi
\ifx \bseriesno  \undefined \def \bseriesno#1{#1}\fi
\ifx \blocation  \undefined \def \blocation#1{#1}\fi
\ifx \bsertitle  \undefined \def \bsertitle#1{#1}\fi
\ifx \bsnm \undefined \def \bsnm#1{#1}\fi
\ifx \bsuffix \undefined \def \bsuffix#1{#1}\fi
\ifx \bparticle \undefined \def \bparticle#1{#1}\fi
\ifx \barticle \undefined \def \barticle#1{#1}\fi
\bibcommenthead
\ifx \bconfdate \undefined \def \bconfdate #1{#1}\fi
\ifx \botherref \undefined \def \botherref #1{#1}\fi
\ifx \url \undefined \def \url#1{\textsf{#1}}\fi
\ifx \bchapter \undefined \def \bchapter#1{#1}\fi
\ifx \bbook \undefined \def \bbook#1{#1}\fi
\ifx \bcomment \undefined \def \bcomment#1{#1}\fi
\ifx \oauthor \undefined \def \oauthor#1{#1}\fi
\ifx \citeauthoryear \undefined \def \citeauthoryear#1{#1}\fi
\ifx \endbibitem  \undefined \def \endbibitem {}\fi
\ifx \bconflocation  \undefined \def \bconflocation#1{#1}\fi
\ifx \arxivurl  \undefined \def \arxivurl#1{\textsf{#1}}\fi
\csname PreBibitemsHook\endcsname

%%% 1
\bibitem[\protect\citeauthoryear{Ghamisi et~al.}{2017}]{ghamisi2017advanced}
\begin{barticle}
\bauthor{\bsnm{Ghamisi}, \binits{P.}},
\bauthor{\bsnm{Plaza}, \binits{J.}},
\bauthor{\bsnm{Chen}, \binits{Y.}},
\bauthor{\bsnm{Li}, \binits{J.}},
\bauthor{\bsnm{Plaza}, \binits{A.J.}}:
\batitle{Advanced spectral classifiers for hyperspectral images: A review}.
\bjtitle{IEEE Geoscience and Remote Sensing Magazine}
\bvolume{5}(\bissue{1}),
\bfpage{8}--\blpage{32}
(\byear{2017})
\end{barticle}
\endbibitem

%%% 2
\bibitem[\protect\citeauthoryear{Xu and Ghamisi}{2022}]{xu2022universal}
\begin{barticle}
\bauthor{\bsnm{Xu}, \binits{Y.}},
\bauthor{\bsnm{Ghamisi}, \binits{P.}}:
\batitle{Universal adversarial examples in remote sensing: Methodology and benchmark}.
\bjtitle{IEEE Transactions on Geoscience and Remote Sensing}
\bvolume{60},
\bfpage{1}--\blpage{15}
(\byear{2022})
\end{barticle}
\endbibitem

%%% 3
\bibitem[\protect\citeauthoryear{Zhang and Zhang}{2022}]{zhang2022artificial}
\begin{barticle}
\bauthor{\bsnm{Zhang}, \binits{L.}},
\bauthor{\bsnm{Zhang}, \binits{L.}}:
\batitle{Artificial intelligence for remote sensing data analysis: A review of challenges and opportunities}.
\bjtitle{IEEE Geoscience and Remote Sensing Magazine}
\bvolume{10}(\bissue{2}),
\bfpage{270}--\blpage{294}
(\byear{2022})
\end{barticle}
\endbibitem

%%% 4
\bibitem[\protect\citeauthoryear{Sermanet et~al.}{2012}]{sermanet2012convolutional}
\begin{bchapter}
\bauthor{\bsnm{Sermanet}, \binits{P.}},
\bauthor{\bsnm{Chintala}, \binits{S.}},
\bauthor{\bsnm{LeCun}, \binits{Y.}}:
\bctitle{Convolutional neural networks applied to house numbers digit classification}.
In: \bbtitle{Proceedings of the 21st International Conference on Pattern Recognition (ICPR2012)},
pp. \bfpage{3288}--\blpage{3291}
(\byear{2012}).
\bcomment{IEEE}
\end{bchapter}
\endbibitem

%%% 5
\bibitem[\protect\citeauthoryear{Goodfellow et~al.}{2014}]{goodfellow2014generative}
\begin{botherref}
\oauthor{\bsnm{Goodfellow}, \binits{I.}},
\oauthor{\bsnm{Pouget-Abadie}, \binits{J.}},
\oauthor{\bsnm{Mirza}, \binits{M.}},
\oauthor{\bsnm{Xu}, \binits{B.}},
\oauthor{\bsnm{Warde-Farley}, \binits{D.}},
\oauthor{\bsnm{Ozair}, \binits{S.}},
\oauthor{\bsnm{Courville}, \binits{A.}},
\oauthor{\bsnm{Bengio}, \binits{Y.}}:
Generative adversarial nets.
Advances in neural information processing systems
\textbf{27}
(2014)
\end{botherref}
\endbibitem

%%% 6
\bibitem[\protect\citeauthoryear{Chen et~al.}{2021}]{chen2021remote}
\begin{barticle}
\bauthor{\bsnm{Chen}, \binits{C.}},
\bauthor{\bsnm{Ma}, \binits{H.}},
\bauthor{\bsnm{Yao}, \binits{G.}},
\bauthor{\bsnm{Lv}, \binits{N.}},
\bauthor{\bsnm{Yang}, \binits{H.}},
\bauthor{\bsnm{Li}, \binits{C.}},
\bauthor{\bsnm{Wan}, \binits{S.}}:
\batitle{Remote sensing image augmentation based on text description for waterside change detection}.
\bjtitle{Remote Sensing}
\bvolume{13}(\bissue{10}),
\bfpage{1894}
(\byear{2021})
\end{barticle}
\endbibitem

%%% 7
\bibitem[\protect\citeauthoryear{Bejiga et~al.}{2019}]{bejiga2019retro}
\begin{barticle}
\bauthor{\bsnm{Bejiga}, \binits{M.B.}},
\bauthor{\bsnm{Melgani}, \binits{F.}},
\bauthor{\bsnm{Vascotto}, \binits{A.}}:
\batitle{Retro-remote sensing: Generating images from ancient texts}.
\bjtitle{IEEE Journal of Selected Topics in Applied Earth Observations and Remote Sensing}
\bvolume{12}(\bissue{3}),
\bfpage{950}--\blpage{960}
(\byear{2019})
\end{barticle}
\endbibitem

%%% 8
\bibitem[\protect\citeauthoryear{Zhao and Shi}{2021}]{zhao2021text}
\begin{barticle}
\bauthor{\bsnm{Zhao}, \binits{R.}},
\bauthor{\bsnm{Shi}, \binits{Z.}}:
\batitle{Text-to-remote-sensing-image generation with structured generative adversarial networks}.
\bjtitle{IEEE Geoscience and Remote Sensing Letters}
\bvolume{19},
\bfpage{1}--\blpage{5}
(\byear{2021})
\end{barticle}
\endbibitem

%%% 9
\bibitem[\protect\citeauthoryear{Ho et~al.}{2020}]{ho2020denoising}
\begin{barticle}
\bauthor{\bsnm{Ho}, \binits{J.}},
\bauthor{\bsnm{Jain}, \binits{A.}},
\bauthor{\bsnm{Abbeel}, \binits{P.}}:
\batitle{Denoising diffusion probabilistic models}.
\bjtitle{Advances in neural information processing systems}
\bvolume{33},
\bfpage{6840}--\blpage{6851}
(\byear{2020})
\end{barticle}
\endbibitem

%%% 10
\bibitem[\protect\citeauthoryear{Ho et~al.}{2022}]{ho2022cascaded}
\begin{barticle}
\bauthor{\bsnm{Ho}, \binits{J.}},
\bauthor{\bsnm{Saharia}, \binits{C.}},
\bauthor{\bsnm{Chan}, \binits{W.}},
\bauthor{\bsnm{Fleet}, \binits{D.J.}},
\bauthor{\bsnm{Norouzi}, \binits{M.}},
\bauthor{\bsnm{Salimans}, \binits{T.}}:
\batitle{Cascaded diffusion models for high fidelity image generation}.
\bjtitle{The Journal of Machine Learning Research}
\bvolume{23}(\bissue{1}),
\bfpage{2249}--\blpage{2281}
(\byear{2022})
\end{barticle}
\endbibitem

%%% 11
\bibitem[\protect\citeauthoryear{Reed et~al.}{2016}]{reed2016learning}
\begin{botherref}
\oauthor{\bsnm{Reed}, \binits{S.E.}},
\oauthor{\bsnm{Akata}, \binits{Z.}},
\oauthor{\bsnm{Mohan}, \binits{S.}},
\oauthor{\bsnm{Tenka}, \binits{S.}},
\oauthor{\bsnm{Schiele}, \binits{B.}},
\oauthor{\bsnm{Lee}, \binits{H.}}:
Learning what and where to draw.
Advances in neural information processing systems
\textbf{29}
(2016)
\end{botherref}
\endbibitem

%%% 12
\bibitem[\protect\citeauthoryear{Zhang et~al.}{2017}]{zhang2017stackgan}
\begin{bchapter}
\bauthor{\bsnm{Zhang}, \binits{H.}},
\bauthor{\bsnm{Xu}, \binits{T.}},
\bauthor{\bsnm{Li}, \binits{H.}},
\bauthor{\bsnm{Zhang}, \binits{S.}},
\bauthor{\bsnm{Wang}, \binits{X.}},
\bauthor{\bsnm{Huang}, \binits{X.}},
\bauthor{\bsnm{Metaxas}, \binits{D.N.}}:
\bctitle{Stackgan: Text to photo-realistic image synthesis with stacked generative adversarial networks}.
In: \bbtitle{Proceedings of the IEEE International Conference on Computer Vision},
pp. \bfpage{5907}--\blpage{5915}
(\byear{2017})
\end{bchapter}
\endbibitem

%%% 13
\bibitem[\protect\citeauthoryear{Brown et~al.}{2020}]{brown2020language}
\begin{barticle}
\bauthor{\bsnm{Brown}, \binits{T.}},
\bauthor{\bsnm{Mann}, \binits{B.}},
\bauthor{\bsnm{Ryder}, \binits{N.}},
\bauthor{\bsnm{Subbiah}, \binits{M.}},
\bauthor{\bsnm{Kaplan}, \binits{J.D.}},
\bauthor{\bsnm{Dhariwal}, \binits{P.}},
\bauthor{\bsnm{Neelakantan}, \binits{A.}},
\bauthor{\bsnm{Shyam}, \binits{P.}},
\bauthor{\bsnm{Sastry}, \binits{G.}},
\bauthor{\bsnm{Askell}, \binits{A.}}, \betal:
\batitle{Language models are few-shot learners}.
\bjtitle{Advances in neural information processing systems}
\bvolume{33},
\bfpage{1877}--\blpage{1901}
(\byear{2020})
\end{barticle}
\endbibitem

%%% 14
\bibitem[\protect\citeauthoryear{Ramesh et~al.}{2021}]{ramesh2021zero}
\begin{bchapter}
\bauthor{\bsnm{Ramesh}, \binits{A.}},
\bauthor{\bsnm{Pavlov}, \binits{M.}},
\bauthor{\bsnm{Goh}, \binits{G.}},
\bauthor{\bsnm{Gray}, \binits{S.}},
\bauthor{\bsnm{Voss}, \binits{C.}},
\bauthor{\bsnm{Radford}, \binits{A.}},
\bauthor{\bsnm{Chen}, \binits{M.}},
\bauthor{\bsnm{Sutskever}, \binits{I.}}:
\bctitle{Zero-shot text-to-image generation}.
In: \bbtitle{International Conference on Machine Learning},
pp. \bfpage{8821}--\blpage{8831}
(\byear{2021}).
\bcomment{PMLR}
\end{bchapter}
\endbibitem

%%% 15
\bibitem[\protect\citeauthoryear{Ramesh et~al.}{2022}]{ramesh2022hierarchical}
\begin{barticle}
\bauthor{\bsnm{Ramesh}, \binits{A.}},
\bauthor{\bsnm{Dhariwal}, \binits{P.}},
\bauthor{\bsnm{Nichol}, \binits{A.}},
\bauthor{\bsnm{Chu}, \binits{C.}},
\bauthor{\bsnm{Chen}, \binits{M.}}:
\batitle{Hierarchical text-conditional image generation with clip latents}.
\bjtitle{arXiv preprint arXiv:2204.06125}
\bvolume{1}(\bissue{2}),
\bfpage{3}
(\byear{2022})
\end{barticle}
\endbibitem

%%% 16
\bibitem[\protect\citeauthoryear{Radford et~al.}{2021}]{radford2021learning}
\begin{bchapter}
\bauthor{\bsnm{Radford}, \binits{A.}},
\bauthor{\bsnm{Kim}, \binits{J.W.}},
\bauthor{\bsnm{Hallacy}, \binits{C.}},
\bauthor{\bsnm{Ramesh}, \binits{A.}},
\bauthor{\bsnm{Goh}, \binits{G.}},
\bauthor{\bsnm{Agarwal}, \binits{S.}},
\bauthor{\bsnm{Sastry}, \binits{G.}},
\bauthor{\bsnm{Askell}, \binits{A.}},
\bauthor{\bsnm{Mishkin}, \binits{P.}},
\bauthor{\bsnm{Clark}, \binits{J.}}, \betal:
\bctitle{Learning transferable visual models from natural language supervision}.
In: \bbtitle{International Conference on Machine Learning},
pp. \bfpage{8748}--\blpage{8763}
(\byear{2021}).
\bcomment{PMLR}
\end{bchapter}
\endbibitem

%%% 17
\bibitem[\protect\citeauthoryear{Chen et~al.}{2024}]{chen2024railfod23}
\begin{barticle}
\bauthor{\bsnm{Chen}, \binits{Z.}},
\bauthor{\bsnm{Yang}, \binits{J.}},
\bauthor{\bsnm{Feng}, \binits{Z.}},
\bauthor{\bsnm{Zhu}, \binits{H.}}:
\batitle{Railfod23: A dataset for foreign object detection on railroad transmission lines}.
\bjtitle{Scientific Data}
\bvolume{11}(\bissue{1}),
\bfpage{72}
(\byear{2024})
\end{barticle}
\endbibitem

%%% 18
\bibitem[\protect\citeauthoryear{Yang et~al.}{2023}]{10035427}
\begin{barticle}
\bauthor{\bsnm{Yang}, \binits{L.}},
\bauthor{\bsnm{Li}, \binits{X.}},
\bauthor{\bsnm{Sun}, \binits{M.}},
\bauthor{\bsnm{Sun}, \binits{C.}}:
\batitle{Hybrid policy-based reinforcement learning of adaptive energy management for the energy transmission-constrained island group}.
\bjtitle{IEEE Transactions on Industrial Informatics}
\bvolume{19}(\bissue{11}),
\bfpage{10751}--\blpage{10762}
(\byear{2023})
\doiurl{10.1109/TII.2023.3241682}
\end{barticle}
\endbibitem

%%% 19
\bibitem[\protect\citeauthoryear{Cui et~al.}{2020}]{8962207}
\begin{barticle}
\bauthor{\bsnm{Cui}, \binits{Y.}},
\bauthor{\bsnm{Wu}, \binits{D.}},
\bauthor{\bsnm{Huang}, \binits{J.}}:
\batitle{Optimize tsk fuzzy systems for classification problems: Minibatch gradient descent with uniform regularization and batch normalization}.
\bjtitle{IEEE Transactions on Fuzzy Systems}
\bvolume{28}(\bissue{12}),
\bfpage{3065}--\blpage{3075}
(\byear{2020})
\doiurl{10.1109/TFUZZ.2020.2967282}
\end{barticle}
\endbibitem

%%% 20
\bibitem[\protect\citeauthoryear{Zhang et~al.}{2024}]{10538013}
\begin{botherref}
\oauthor{\bsnm{Zhang}, \binits{N.}},
\oauthor{\bsnm{Yan}, \binits{J.}},
\oauthor{\bsnm{Hu}, \binits{C.}},
\oauthor{\bsnm{Sun}, \binits{Q.}},
\oauthor{\bsnm{Yang}, \binits{L.}},
\oauthor{\bsnm{Gao}, \binits{D.W.}},
\oauthor{\bsnm{Guerrero}, \binits{J.M.}},
\oauthor{\bsnm{Li}, \binits{Y.}}:
Price-matching-based regional energy market with hierarchical reinforcement learning algorithm.
IEEE Transactions on Industrial Informatics,
1--12
(2024)
\doiurl{10.1109/TII.2024.3390595}
\end{botherref}
\endbibitem

%%% 21
\bibitem[\protect\citeauthoryear{Li et~al.}{2019}]{8424453}
\begin{barticle}
\bauthor{\bsnm{Li}, \binits{Y.}},
\bauthor{\bsnm{Zhang}, \binits{H.}},
\bauthor{\bsnm{Liang}, \binits{X.}},
\bauthor{\bsnm{Huang}, \binits{B.}}:
\batitle{Event-triggered-based distributed cooperative energy management for multienergy systems}.
\bjtitle{IEEE Transactions on Industrial Informatics}
\bvolume{15}(\bissue{4}),
\bfpage{2008}--\blpage{2022}
(\byear{2019})
\doiurl{10.1109/TII.2018.2862436}
\end{barticle}
\endbibitem

%%% 22
\bibitem[\protect\citeauthoryear{Saharia et~al.}{2022}]{saharia2022photorealistic}
\begin{barticle}
\bauthor{\bsnm{Saharia}, \binits{C.}},
\bauthor{\bsnm{Chan}, \binits{W.}},
\bauthor{\bsnm{Saxena}, \binits{S.}},
\bauthor{\bsnm{Li}, \binits{L.}},
\bauthor{\bsnm{Whang}, \binits{J.}},
\bauthor{\bsnm{Denton}, \binits{E.L.}},
\bauthor{\bsnm{Ghasemipour}, \binits{K.}},
\bauthor{\bsnm{Gontijo~Lopes}, \binits{R.}},
\bauthor{\bsnm{Karagol~Ayan}, \binits{B.}},
\bauthor{\bsnm{Salimans}, \binits{T.}}, \betal:
\batitle{Photorealistic text-to-image diffusion models with deep language understanding}.
\bjtitle{Advances in Neural Information Processing Systems}
\bvolume{35},
\bfpage{36479}--\blpage{36494}
(\byear{2022})
\end{barticle}
\endbibitem

%%% 23
\bibitem[\protect\citeauthoryear{Devlin et~al.}{2018}]{devlin2018bert}
\begin{botherref}
\oauthor{\bsnm{Devlin}, \binits{J.}},
\oauthor{\bsnm{Chang}, \binits{M.-W.}},
\oauthor{\bsnm{Lee}, \binits{K.}},
\oauthor{\bsnm{Toutanova}, \binits{K.}}:
Bert: Pre-training of deep bidirectional transformers for language understanding.
arXiv preprint arXiv:1810.04805
(2018)
\end{botherref}
\endbibitem

%%% 24
\bibitem[\protect\citeauthoryear{Raffel et~al.}{2020}]{raffel2020exploring}
\begin{barticle}
\bauthor{\bsnm{Raffel}, \binits{C.}},
\bauthor{\bsnm{Shazeer}, \binits{N.}},
\bauthor{\bsnm{Roberts}, \binits{A.}},
\bauthor{\bsnm{Lee}, \binits{K.}},
\bauthor{\bsnm{Narang}, \binits{S.}},
\bauthor{\bsnm{Matena}, \binits{M.}},
\bauthor{\bsnm{Zhou}, \binits{Y.}},
\bauthor{\bsnm{Li}, \binits{W.}},
\bauthor{\bsnm{Liu}, \binits{P.J.}}:
\batitle{Exploring the limits of transfer learning with a unified text-to-text transformer}.
\bjtitle{The Journal of Machine Learning Research}
\bvolume{21}(\bissue{1}),
\bfpage{5485}--\blpage{5551}
(\byear{2020})
\end{barticle}
\endbibitem

%%% 25
\bibitem[\protect\citeauthoryear{Raffel et~al.}{2017}]{raffel2017online}
\begin{bchapter}
\bauthor{\bsnm{Raffel}, \binits{C.}},
\bauthor{\bsnm{Luong}, \binits{M.-T.}},
\bauthor{\bsnm{Liu}, \binits{P.J.}},
\bauthor{\bsnm{Weiss}, \binits{R.J.}},
\bauthor{\bsnm{Eck}, \binits{D.}}:
\bctitle{Online and linear-time attention by enforcing monotonic alignments}.
In: \bbtitle{International Conference on Machine Learning},
pp. \bfpage{2837}--\blpage{2846}
(\byear{2017}).
\bcomment{PMLR}
\end{bchapter}
\endbibitem

%%% 26
\bibitem[\protect\citeauthoryear{Sohl-Dickstein et~al.}{2015}]{sohl2015deep}
\begin{bchapter}
\bauthor{\bsnm{Sohl-Dickstein}, \binits{J.}},
\bauthor{\bsnm{Weiss}, \binits{E.}},
\bauthor{\bsnm{Maheswaranathan}, \binits{N.}},
\bauthor{\bsnm{Ganguli}, \binits{S.}}:
\bctitle{Deep unsupervised learning using nonequilibrium thermodynamics}.
In: \bbtitle{International Conference on Machine Learning},
pp. \bfpage{2256}--\blpage{2265}
(\byear{2015}).
\bcomment{PMLR}
\end{bchapter}
\endbibitem

%%% 27
\bibitem[\protect\citeauthoryear{Song and Ermon}{2019}]{song2019generative}
\begin{botherref}
\oauthor{\bsnm{Song}, \binits{Y.}},
\oauthor{\bsnm{Ermon}, \binits{S.}}:
Generative modeling by estimating gradients of the data distribution.
Advances in neural information processing systems
\textbf{32}
(2019)
\end{botherref}
\endbibitem

%%% 28
\bibitem[\protect\citeauthoryear{Dhariwal and Nichol}{2021}]{dhariwal2021diffusion}
\begin{barticle}
\bauthor{\bsnm{Dhariwal}, \binits{P.}},
\bauthor{\bsnm{Nichol}, \binits{A.}}:
\batitle{Diffusion models beat gans on image synthesis}.
\bjtitle{Advances in neural information processing systems}
\bvolume{34},
\bfpage{8780}--\blpage{8794}
(\byear{2021})
\end{barticle}
\endbibitem

%%% 29
\bibitem[\protect\citeauthoryear{Nichol et~al.}{2021}]{nichol2021glide}
\begin{botherref}
\oauthor{\bsnm{Nichol}, \binits{A.}},
\oauthor{\bsnm{Dhariwal}, \binits{P.}},
\oauthor{\bsnm{Ramesh}, \binits{A.}},
\oauthor{\bsnm{Shyam}, \binits{P.}},
\oauthor{\bsnm{Mishkin}, \binits{P.}},
\oauthor{\bsnm{McGrew}, \binits{B.}},
\oauthor{\bsnm{Sutskever}, \binits{I.}},
\oauthor{\bsnm{Chen}, \binits{M.}}:
Glide: Towards photorealistic image generation and editing with text-guided diffusion models.
arXiv preprint arXiv:2112.10741
(2021)
\end{botherref}
\endbibitem

%%% 30
\bibitem[\protect\citeauthoryear{Saharia et~al.}{2022a}]{saharia2022palette}
\begin{bchapter}
\bauthor{\bsnm{Saharia}, \binits{C.}},
\bauthor{\bsnm{Chan}, \binits{W.}},
\bauthor{\bsnm{Chang}, \binits{H.}},
\bauthor{\bsnm{Lee}, \binits{C.}},
\bauthor{\bsnm{Ho}, \binits{J.}},
\bauthor{\bsnm{Salimans}, \binits{T.}},
\bauthor{\bsnm{Fleet}, \binits{D.}},
\bauthor{\bsnm{Norouzi}, \binits{M.}}:
\bctitle{Palette: Image-to-image diffusion models}.
In: \bbtitle{ACM SIGGRAPH 2022 Conference Proceedings},
pp. \bfpage{1}--\blpage{10}
(\byear{2022})
\end{bchapter}
\endbibitem

%%% 31
\bibitem[\protect\citeauthoryear{Saharia et~al.}{2022b}]{saharia2022image}
\begin{barticle}
\bauthor{\bsnm{Saharia}, \binits{C.}},
\bauthor{\bsnm{Ho}, \binits{J.}},
\bauthor{\bsnm{Chan}, \binits{W.}},
\bauthor{\bsnm{Salimans}, \binits{T.}},
\bauthor{\bsnm{Fleet}, \binits{D.J.}},
\bauthor{\bsnm{Norouzi}, \binits{M.}}:
\batitle{Image super-resolution via iterative refinement}.
\bjtitle{IEEE Transactions on Pattern Analysis and Machine Intelligence}
\bvolume{45}(\bissue{4}),
\bfpage{4713}--\blpage{4726}
(\byear{2022})
\end{barticle}
\endbibitem

%%% 32
\bibitem[\protect\citeauthoryear{Whang et~al.}{2022}]{whang2022deblurring}
\begin{bchapter}
\bauthor{\bsnm{Whang}, \binits{J.}},
\bauthor{\bsnm{Delbracio}, \binits{M.}},
\bauthor{\bsnm{Talebi}, \binits{H.}},
\bauthor{\bsnm{Saharia}, \binits{C.}},
\bauthor{\bsnm{Dimakis}, \binits{A.G.}},
\bauthor{\bsnm{Milanfar}, \binits{P.}}:
\bctitle{Deblurring via stochastic refinement}.
In: \bbtitle{Proceedings of the IEEE/CVF Conference on Computer Vision and Pattern Recognition},
pp. \bfpage{16293}--\blpage{16303}
(\byear{2022})
\end{bchapter}
\endbibitem

%%% 33
\bibitem[\protect\citeauthoryear{Ho and Salimans}{2022}]{ho2022classifier}
\begin{botherref}
\oauthor{\bsnm{Ho}, \binits{J.}},
\oauthor{\bsnm{Salimans}, \binits{T.}}:
Classifier-free diffusion guidance.
arXiv preprint arXiv:2207.12598
(2022)
\end{botherref}
\endbibitem

%%% 34
\bibitem[\protect\citeauthoryear{Nichol and Dhariwal}{2021}]{nichol2021improved}
\begin{bchapter}
\bauthor{\bsnm{Nichol}, \binits{A.Q.}},
\bauthor{\bsnm{Dhariwal}, \binits{P.}}:
\bctitle{Improved denoising diffusion probabilistic models}.
In: \bbtitle{International Conference on Machine Learning},
pp. \bfpage{8162}--\blpage{8171}
(\byear{2021}).
\bcomment{PMLR}
\end{bchapter}
\endbibitem

%%% 35
\bibitem[\protect\citeauthoryear{Song et~al.}{2020}]{song2020score}
\begin{botherref}
\oauthor{\bsnm{Song}, \binits{Y.}},
\oauthor{\bsnm{Sohl-Dickstein}, \binits{J.}},
\oauthor{\bsnm{Kingma}, \binits{D.P.}},
\oauthor{\bsnm{Kumar}, \binits{A.}},
\oauthor{\bsnm{Ermon}, \binits{S.}},
\oauthor{\bsnm{Poole}, \binits{B.}}:
Score-based generative modeling through stochastic differential equations.
arXiv preprint arXiv:2011.13456
(2020)
\end{botherref}
\endbibitem

%%% 36
\bibitem[\protect\citeauthoryear{Lu et~al.}{2017}]{lu2017exploring}
\begin{barticle}
\bauthor{\bsnm{Lu}, \binits{X.}},
\bauthor{\bsnm{Wang}, \binits{B.}},
\bauthor{\bsnm{Zheng}, \binits{X.}},
\bauthor{\bsnm{Li}, \binits{X.}}:
\batitle{Exploring models and data for remote sensing image caption generation}.
\bjtitle{IEEE Transactions on Geoscience and Remote Sensing}
\bvolume{56}(\bissue{4}),
\bfpage{2183}--\blpage{2195}
(\byear{2017})
\end{barticle}
\endbibitem

%%% 37
\bibitem[\protect\citeauthoryear{Xu et~al.}{2022}]{xu2022txt2img}
\begin{botherref}
\oauthor{\bsnm{Xu}, \binits{Y.}},
\oauthor{\bsnm{Yu}, \binits{W.}},
\oauthor{\bsnm{Ghamisi}, \binits{P.}},
\oauthor{\bsnm{Kopp}, \binits{M.}},
\oauthor{\bsnm{Hochreiter}, \binits{S.}}:
Txt2img-mhn: Remote sensing image generation from text using modern hopfield networks.
arXiv preprint arXiv:2208.04441
(2022)
\end{botherref}
\endbibitem

%%% 38
\bibitem[\protect\citeauthoryear{Salimans et~al.}{2016}]{salimans2016improved}
\begin{botherref}
\oauthor{\bsnm{Salimans}, \binits{T.}},
\oauthor{\bsnm{Goodfellow}, \binits{I.}},
\oauthor{\bsnm{Zaremba}, \binits{W.}},
\oauthor{\bsnm{Cheung}, \binits{V.}},
\oauthor{\bsnm{Radford}, \binits{A.}},
\oauthor{\bsnm{Chen}, \binits{X.}}:
Improved techniques for training gans.
Advances in neural information processing systems
\textbf{29}
(2016)
\end{botherref}
\endbibitem

%%% 39
\bibitem[\protect\citeauthoryear{Heusel et~al.}{2017}]{heusel2017gans}
\begin{botherref}
\oauthor{\bsnm{Heusel}, \binits{M.}},
\oauthor{\bsnm{Ramsauer}, \binits{H.}},
\oauthor{\bsnm{Unterthiner}, \binits{T.}},
\oauthor{\bsnm{Nessler}, \binits{B.}},
\oauthor{\bsnm{Hochreiter}, \binits{S.}}:
Gans trained by a two time-scale update rule converge to a local nash equilibrium.
Advances in neural information processing systems
\textbf{30}
(2017)
\end{botherref}
\endbibitem

%%% 40
\bibitem[\protect\citeauthoryear{Deng et~al.}{2009}]{deng2009imagenet}
\begin{bchapter}
\bauthor{\bsnm{Deng}, \binits{J.}},
\bauthor{\bsnm{Dong}, \binits{W.}},
\bauthor{\bsnm{Socher}, \binits{R.}},
\bauthor{\bsnm{Li}, \binits{L.-J.}},
\bauthor{\bsnm{Li}, \binits{K.}},
\bauthor{\bsnm{Fei-Fei}, \binits{L.}}:
\bctitle{Imagenet: A large-scale hierarchical image database}.
In: \bbtitle{2009 IEEE Conference on Computer Vision and Pattern Recognition},
pp. \bfpage{248}--\blpage{255}
(\byear{2009}).
\bcomment{Ieee}
\end{bchapter}
\endbibitem

%%% 41
\bibitem[\protect\citeauthoryear{Barratt and Sharma}{2018}]{barratt2018note}
\begin{botherref}
\oauthor{\bsnm{Barratt}, \binits{S.}},
\oauthor{\bsnm{Sharma}, \binits{R.}}:
A note on the inception score.
arXiv preprint arXiv:1801.01973
(2018)
\end{botherref}
\endbibitem

%%% 42
\bibitem[\protect\citeauthoryear{Szegedy et~al.}{2016}]{szegedy2016rethinking}
\begin{bchapter}
\bauthor{\bsnm{Szegedy}, \binits{C.}},
\bauthor{\bsnm{Vanhoucke}, \binits{V.}},
\bauthor{\bsnm{Ioffe}, \binits{S.}},
\bauthor{\bsnm{Shlens}, \binits{J.}},
\bauthor{\bsnm{Wojna}, \binits{Z.}}:
\bctitle{Rethinking the inception architecture for computer vision}.
In: \bbtitle{Proceedings of the IEEE Conference on Computer Vision and Pattern Recognition},
pp. \bfpage{2818}--\blpage{2826}
(\byear{2016})
\end{bchapter}
\endbibitem

%%% 43
\bibitem[\protect\citeauthoryear{Zhou et~al.}{}]{zhou2111lafite}
\begin{botherref}
\oauthor{\bsnm{Zhou}, \binits{Y.}},
\oauthor{\bsnm{Zhang}, \binits{R.}},
\oauthor{\bsnm{Chen}, \binits{C.}},
\oauthor{\bsnm{Li}, \binits{C.}},
\oauthor{\bsnm{Tensmeyer}, \binits{C.}},
\oauthor{\bsnm{Yu}, \binits{T.}},
\oauthor{\bsnm{Gu}, \binits{J.}},
\oauthor{\bsnm{Xu}, \binits{J.}},
\oauthor{\bsnm{Sun}, \binits{T.}}:
Lafite: Towards language-free training for text-to-image generation. arxiv 2021.
arXiv preprint arXiv:2111.13792
\end{botherref}
\endbibitem

%%% 44
\bibitem[\protect\citeauthoryear{Shazeer and Stern}{2018}]{shazeer2018adafactor}
\begin{bchapter}
\bauthor{\bsnm{Shazeer}, \binits{N.}},
\bauthor{\bsnm{Stern}, \binits{M.}}:
\bctitle{Adafactor: Adaptive learning rates with sublinear memory cost}.
In: \bbtitle{International Conference on Machine Learning},
pp. \bfpage{4596}--\blpage{4604}
(\byear{2018}).
\bcomment{PMLR}
\end{bchapter}
\endbibitem

%%% 45
\bibitem[\protect\citeauthoryear{Kingma and Ba}{2014}]{kingma2014adam}
\begin{botherref}
\oauthor{\bsnm{Kingma}, \binits{D.P.}},
\oauthor{\bsnm{Ba}, \binits{J.}}:
Adam: A method for stochastic optimization.
arXiv preprint arXiv:1412.6980
(2014)
\end{botherref}
\endbibitem

%%% 46
\bibitem[\protect\citeauthoryear{Xu et~al.}{2018}]{xu2018attngan}
\begin{bchapter}
\bauthor{\bsnm{Xu}, \binits{T.}},
\bauthor{\bsnm{Zhang}, \binits{P.}},
\bauthor{\bsnm{Huang}, \binits{Q.}},
\bauthor{\bsnm{Zhang}, \binits{H.}},
\bauthor{\bsnm{Gan}, \binits{Z.}},
\bauthor{\bsnm{Huang}, \binits{X.}},
\bauthor{\bsnm{He}, \binits{X.}}:
\bctitle{Attngan: Fine-grained text to image generation with attentional generative adversarial networks}.
In: \bbtitle{Proceedings of the IEEE Conference on Computer Vision and Pattern Recognition},
pp. \bfpage{1316}--\blpage{1324}
(\byear{2018})
\end{bchapter}
\endbibitem

%%% 47
\bibitem[\protect\citeauthoryear{Ruan et~al.}{}]{ruandae}
\begin{botherref}
\oauthor{\bsnm{Ruan}, \binits{S.}},
\oauthor{\bsnm{Zhang}, \binits{Y.}},
\oauthor{\bsnm{Zhang}, \binits{K.}},
\oauthor{\bsnm{Fan}, \binits{Y.}},
\oauthor{\bsnm{Tang}, \binits{F.}},
\oauthor{\bsnm{Liu}, \binits{Q.}},
\oauthor{\bsnm{Chen}, \binits{E.}}:
Dae-gan: Dynamic aspect-aware gan for text-to-image synthesis supplementary document
\end{botherref}
\endbibitem

%%% 48
\bibitem[\protect\citeauthoryear{Tao et~al.}{2022}]{tao2022df}
\begin{bchapter}
\bauthor{\bsnm{Tao}, \binits{M.}},
\bauthor{\bsnm{Tang}, \binits{H.}},
\bauthor{\bsnm{Wu}, \binits{F.}},
\bauthor{\bsnm{Jing}, \binits{X.-Y.}},
\bauthor{\bsnm{Bao}, \binits{B.-K.}},
\bauthor{\bsnm{Xu}, \binits{C.}}:
\bctitle{Df-gan: A simple and effective baseline for text-to-image synthesis}.
In: \bbtitle{Proceedings of the IEEE/CVF Conference on Computer Vision and Pattern Recognition},
pp. \bfpage{16515}--\blpage{16525}
(\byear{2022})
\end{bchapter}
\endbibitem

\end{thebibliography}
%% if required, the content of .bbl file can be included here once bbl is generated
%%\input sn-article.bbl

\end{document}